\documentclass{article}

 \PassOptionsToPackage{numbers}{natbib}
    \usepackage[final]{neurips_data_2021}

\usepackage[utf8]{inputenc} % allow utf-8 input
\usepackage[T1]{fontenc}    % use 8-bit T1 fonts
\usepackage{hyperref}       % hyperlinks
\usepackage{url}            % simple URL typesetting
\usepackage{booktabs}       % professional-quality tables
\usepackage{amsfonts}       % blackboard math symbols
\usepackage{nicefrac}       % compact symbols for 1/2, etc.
\usepackage{microtype}      % microtypography
\usepackage{xcolor}         % colors

\usepackage{amsmath} % for align
\usepackage{makecell} % for linebreak in table
\usepackage{colortbl} % for table row color
\usepackage{multirow} % for merging multiple vertical cells in table
\usepackage{url}
\newcommand{\argmax}{\operatornamewithlimits{argmax}} % for argmax in math
\usepackage{subcaption}
\usepackage{graphicx}
\usepackage{listings} % for code
\usepackage{xcolor}

\definecolor{codegreen}{rgb}{0,0.6,0}
\definecolor{codegray}{rgb}{0.5,0.5,0.5}
\definecolor{codepurple}{rgb}{0.58,0,0.82}
\definecolor{backcolour}{rgb}{0.95,0.95,0.92}

\usepackage{inconsolata}
\lstdefinestyle{mystyle}{
    backgroundcolor=\color{backcolour},   
    commentstyle=\color{codegreen},
    keywordstyle=\color{magenta},
    numberstyle=\tiny\color{codegray},
    stringstyle=\color{codepurple},
    basicstyle=\ttfamily\footnotesize,
    breakatwhitespace=false,         
    breaklines=true,                 
    captionpos=b,                    
    keepspaces=true,                 
    numbers=left,                    
    numbersep=5pt,                  
    showspaces=false,                
    showstringspaces=false,
    xleftmargin=14pt,
    framexleftmargin=14pt,
    showtabs=false,                  
    tabsize=2
}

\title{Timers and Such: A Practical Benchmark for Spoken Language Understanding with Numbers}

\author{%
  Loren Lugosch \\
  McGill University / Mila\\
  \texttt{lugoschl@mila.quebec} \\
   \And
Piyush Papreja \\
   \texttt{ppapreja@asu.edu} \\
   \AND
   Mirco Ravanelli \\
   Université de Montréal / Mila\\
   \texttt{mirco.ravanelli@gmail.com} \\
   \And
   Abdelwahab Heba \\
   Université Paul Sabatier, IRIT, CNRS \\
   \texttt{aheba@irit.fr} \\
   \And
   Titouan Parcollet \\
   Avignon Université \\
   \texttt{titouan.parcollet@univ-avignon.fr} \\
}

\begin{document}

\maketitle

\begin{abstract}
  This paper introduces Timers and Such, a new open source dataset of spoken English commands for common voice control use cases involving numbers. We describe the gap in existing spoken language understanding datasets that Timers and Such fills, the design and creation of the dataset, and experiments with a number of ASR-based and end-to-end baseline models, the code for which has been made available as part of the SpeechBrain toolkit.
\end{abstract}

\section{Introduction}\label{intro}

Spoken language understanding (SLU) research has begun to emphasize the importance of both \textit{testing} and \textit{training} SLU systems end-to-end on audio. 
\textit{Testing} on audio is important because an independently trained automatic speech recognition (ASR) system and natural language understanding (NLU) system will not necessarily work well when combined \cite{hakkani2006beyond, saade2018spoken}. \textit{Training} SLU systems end-to-end on audio is likewise worthwhile because it can make the NLU model more robust to transcription errors, and because it enables training a single neural network to perform the entire SLU pipeline without an intermediate search step, a technique with many practical and theoretical advantages over ASR-based approaches \cite{Serdyuk2018}. 

Experiments involving end-to-end training and testing of SLU models require audio data. Over the last few years, a number of open source audio datasets have been released to enable high-quality, reproducible end-to-end SLU research. The \textbf{Snips SLU Dataset} \cite{saade2018spoken} is a small dataset of English and French commands for a smart home setting, such as controlling smart lights, speaker volume, and music selection. \textbf{Fluent Speech Commands} \cite{lugosch2019speech} is a somewhat larger, though simpler, dataset of similar English smart home commands. The most recently released \textbf{SLURP} dataset \cite{slurp} is an even larger and much more semantically complex multi-domain SLU dataset. 

An important feature missing from these datasets is a thorough coverage of \textit{numbers}. Numbers are necessary for many SLU domains, especially for very common use cases like setting timers and converting units of measurement while cooking. 
While there do exist datasets of digits spoken in isolation \cite{leonard1993tidigits, warden2018speech, zohar_jackson_2018_1342401}, and the Snips SLU Dataset and SLURP do have a small number of commands involving simple numbers, there does not to our knowledge exist any open source SLU dataset that covers more general multi-digit numbers (e.g.~``13.57'', ``-21.4'') spoken in context. The dataset introduced here---\textbf{Timers and Such}---fills this gap, with each command containing one or two numbers with one or more digits. 

One of the original motivations for the development of end-to-end SLU models was the need for more compact models that can easily fit on resource-limited devices and operate without an Internet connection \cite{Serdyuk2018}. Whereas existing SLU datasets focus mostly on Internet-connected smart home commands or queries that require an Internet search, Timers and Such is composed only of commands that can be executed without the need for the Internet. This makes the dataset ideal for training or testing a simple offline voice assistant. While the baselines described in this paper all use rather comfortably large neural networks ($>$100 million parameters), we hope that researchers and developers working on machine learning for edge devices will improve upon our models in terms of storage requirements and computational complexity; we believe they will find Timers and Such to be a challenging and interesting test case for their models. 

The dataset should also be useful for researchers working on representation learning for audio and language to use as a downstream test task, as Fluent Speech Commands has been \cite{tamkin2020viewmaker, chung2020semi}. While in the past we have found supervised ASR-based pre-training to be essential for getting good results with end-to-end SLU models, we believe unsupervised feature extractors may ultimately prove to be a better general-purpose solution for SLU and other audio tasks \cite{pascual2019learning, baevski2020wav2vec}.

A final, more mundane motivation for Timers and Such was the need for an SLU dataset that could easily be downloaded programmatically using tools like \texttt{wget} or \texttt{curl}, similar to MNIST or LibriSpeech.\footnote{SLURP, released after the start of this work, can also be downloaded programmatically.} Fluent Speech Commands requires users to sign up on a web page, and the Snips SLU dataset requires filling in an online form and waiting to be approved. In contrast to these, Timers and Such is hosted on Zenodo\footnote{The dataset can be found at \url{https://zenodo.org/record/4623772}.} under the very permissive CC0 license, and the experiment code\footnote{The code can be found at \url{https://github.com/speechbrain/speechbrain/tree/develop/recipes/timers-and-such}.} we provide downloads the dataset if it is not already present in the location specified by the user. These features should lower the barrier to entry for anyone interested in training or testing their first SLU model.

In what follows, we outline the design and creation of Timers and Such, describe some baseline models for the dataset, discuss their experimental performance, and end by listing some ideas for future research.

\section{Dataset design}

% ("what's 37.67 minus 75.7",
% {
%   'intent': 'SimpleMath', 
%   'slots': {
%     'number1': 37.67, 
%     'number2': 75.7, 
%     'op': ' minus '
%   }
% })

% \begin{figure}
%     \centering
%     \includegraphics[width=.375\linewidth]{example3.png}
%     \caption{A \texttt{SimpleMath} command and its label dictionary.} 
%     \label{examples}
%   \end{figure}
  
\begin{lstlisting}[float,language=Python, frame=lines, basicstyle=\footnotesize\ttfamily, caption={A \texttt{SimpleMath} command and its label dictionary.}, label={examples}]
("what's 37.67 minus 75.7",
{
  'intent': 'SimpleMath', 
  'slots': {
    'number1': 37.67, 
    'number2': 75.7, 
    'op': ' minus '
  }
})
\end{lstlisting}

The dataset has four intents, corresponding to four common offline voice assistant uses: \texttt{SetTimer}, \texttt{SetAlarm}, \texttt{SimpleMath}, and \texttt{UnitConversion}. The semantic label for each utterance is a dictionary with the intent and a number of slots. An example of a command and its corresponding semantics is shown in Listing \ref{examples}.

The prompts to be recorded by speakers were generated using a script written by the first author with a simple ``grammar'' that produced a few variations of set phrases for each of the four intents (``set a timer for\dots'', ``set timer for\dots'', ``start timer for\dots'').
Random numbers were inserted from a range that made sense for the given intent (for instance, when converting temperatures, temperatures less than 0 Kelvin were not used).\footnote{The script for generating prompts can be found at \url{https://gist.github.com/lorenlugosch/5df9e30227aa5c67ff51cd28271414f0}.}

A better way to collect different ways of phrasing commands than introspection is to place speakers in a voice control scenario (or have them imagine themselves in one) and ask them what they would say to have the system complete a certain task. This method was used to create part of the closed source Facebook dataset in \cite{Serdyuk2018} and the open source SLURP \cite{slurp}. However, this approach is complicated to set up and much more taxing on speakers. Given that our speakers were volunteers, we decided instead to simply prompt them with randomly generated phrases for each of the intents, similar to the approach used in Mozilla's Common Voice project \cite{ardila2019common}.

\section{Preliminary small-scale study}
A preliminary version of Timers and Such was made between November 2019 and October 2020. 11 colleagues recorded themselves reading a list of prompts, some using the first author's laptop, and others using their own computers. The first author then segmented these audio files into the individual commands and split the resulting 271 audios into a training set with 144 audios (4 speakers), a dev set with 72 audios (2 speakers), and a test set with 55 audios (5 speakers). Models trained on this small dataset were found to have high variability in performance for the test set, which was hypothesized to be because of the small test set size. (This actually seems not to have been the real reason; see Sec.~\ref{results-text}.) To make a dataset that could be used more reliably to train and compare SLU models, we decided to reach out to a larger pool of speakers by asking volunteers online to donate their voices.

\section{Data collection}\label{data-collection}

\subsection{Recording website}\label{recording-website}
\begin{figure}
    \centering
    \includegraphics[scale=0.4]{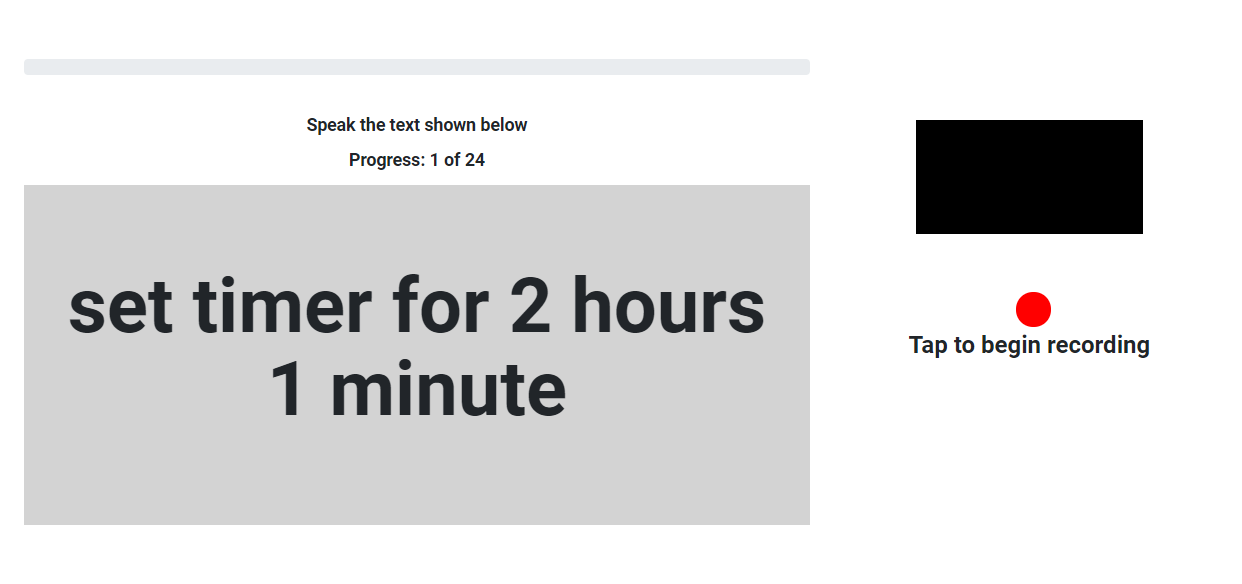}
    \caption{The recording interface used by speakers.}
    \label{fig:recording-website}
\end{figure}

The second author built a website to allow speakers to record themselves reading prompts. Speakers using the website were first asked for their age, gender, and spoken English proficiency. For each demographic field, users also had the option to respond ``Prefer not to say''. After giving their consent to have their demographic information and recordings released in a publicly available dataset, speakers used the interface shown in Fig.~\ref{fig:recording-website} to record a set of 24 randomly generated prompts.

\subsection{Speaker recruitment}\label{speaker-section}
Starting on February 18, 2021, we advertised the project and recording website on various social media platforms (Twitter, LinkedIn, Reddit, Hacker News, Facebook). In response to this advertisement, 89 sessions were recorded from the first day until March 12, 2021.

Whether the 89 recorded sessions correspond to exactly 89 different speakers is unknown. We neglected to ask speakers in the recording instructions not to record more than one session. Because speakers were (deliberately) not asked to provide any information that would uniquely identify them, such as their name or email address, there is no way to ascertain whether two sessions correspond to the same speaker (as is the case for recording platforms like Common Voice's, which allow a speaker to record without entering any personally identifiable information). 
% Hence, a random train-test split on sessions could result in some overlap in speakers between the training set and the validation and test sets, which is considered bad practice when constructing a speech recognition dataset.
To avoid an overlap between speakers in the training set and the test set, we examined the demographic information provided by speakers (age, gender, fluency) and selected only sessions with a unique demographic triple to be in the test set. Assuming speakers provided their demographic information truthfully, this means there are no speakers from the test set in the training set.

\subsection{Data preprocessing and cleaning}\label{data-prep}

All recordings were converted from their original formats to single-channel 16,000 Hz .wav files for compatibility with the ASR model used in our baseline experiments.

Data cleaning for the smaller set of audios collected during the preliminary small-scale study was done manually by the first author. The 271 audios collected in the preliminary study were assigned to the \texttt{dev-real} subset. Those speakers were not asked for their demographic information, so that information is not provided for this split. 

For the larger set of audios recorded using the recording website, we used a more automated form of cleaning: the audios were transcribed using an ASR model (described in Sec.~\ref{ASR}), and the word error rate (WER) between each prompt and transcript was computed. Audios for which the ASR transcript was empty or looked significantly different from the prompt were listened to and kept or deleted as appropriate. 
(A simple automatic decision rule that was found to yield nearly the same subset was to select all audios with WER less than 100\%.) 
After this cleaning procedure, the remaining 1,880 audios were split into \texttt{train-real} and \texttt{test-real} subsets. A .csv file for each subset (\texttt{\{train-real, dev-real, test-real\}.csv}) lists, for each utterance, the .wav filename, the semantic label dictionary, the session ID ($\approx$ speaker ID), and the transcript.

\subsection{Synthetic data}
Following \cite{lugosch2020using}, we used VoiceLoop \cite{taigman2017voice} to synthesize a large set of audios from 22 synthetic speakers. (The VoiceLoop model is trained on the VCTK dataset \cite{veaux2017cstr}.) That set was split by speaker into the \texttt{train-synth} (192,000 audios), \texttt{dev-synth} (24,000 audios), and \texttt{test-synth} (36,000 audios) subsets. As for the data from the real speakers, we include a .csv file (\texttt{\{train-synth, dev-synth, test-synth\}.csv}) listing the filename, semantics, speaker ID (a number 1 to 22 indicating which VoiceLoop synthetic speaker was used), and transcript.
The VoiceLoop speech synthesizer is deterministic: running it on the same prompt twice produces the same audio signal. As a result, some of the rows in the .csv file describing the synthetic subset are redundant: they point to the same audio file with the same labels. We have not removed the redundant rows because we found that doing so led to an unbalanced training set: for example, there were many more instances of ``set alarm for \texttt{<hour>} \texttt{<minute>} AM'' than of ``set alarm for \texttt{<hour>} AM'', so models trained on this unbalanced dataset tended to hallucinate an erroneous value for the \texttt{<minute>} slot for the latter type of utterance. (Alternately, users can rebalance the data in a different way, if they choose, using e.g. \texttt{pandas.DataFrame.drop\_duplicates()} on the filename column of the .csv file.) 
We encourage users of Timers and Such not to think of the synthetic subset as \textit{fixed} (except to avoid unfair comparisons between two models differing in some other respects), but rather to try adding more synthetic speakers and using improved speech synthesis techniques. 

\subsection{Dataset statistics}\label{statistics}
The overall statistics for both the real and synthetic subsets of Timers and Such after data cleaning are listed in Table \ref{stats-counts}. At 2,151 non-synthetic utterances, Timers and Such is a fairly small dataset, but like TIMIT (6,300 utterances \cite{timit}) and the Snips ``smart lights'' dataset (1,660 utterances \cite{saade2018spoken}), we have found the dataset nonetheless very useful for experimentation. It is more challenging than Fluent Speech Commands (which can be treated as a simple classification problem and for which accuracy as high as 99.7\% has been achieved \cite{seo2021integration}), but it is smaller and simpler than SLURP. By training only on text or synthetic speech, and testing on all available real audio, it is possible to obtain a relatively large test set (cf. the LibriSpeech \texttt{test-clean} subset with 2,620 audios).

\begin{table}[ht]
  \caption{Timers and Such speaker counts and recording statistics. ($^*$Speaker counts are approximate; see Section \ref{speaker-section}.)}
  \centering
  \begin{tabular}{l r r r}
    \toprule
    Split & \# of speakers$^*$ & \# of audios & \# hours\\
    \midrule
    \texttt{train-synth} & 16 &  192,000  & 132.2  \\
    \texttt{dev-synth}  & 2 &   24,000  &  15.8  \\
    \texttt{test-synth}  & 3 & 36,000   &  23.5 \\
    \midrule
    \texttt{train-real} & 74 &  1,640  & 1.9  \\
    \texttt{dev-real}  & 11 &   271  &  0.3  \\
    \texttt{test-real}  & 10 & 240   &  0.3 \\
    \midrule
    \texttt{all-real} & 95 & 2,151   &  2.5 \\
    \bottomrule
  \end{tabular}\label{stats-counts}
  
\end{table}

\begin{table}[ht]
  \caption{Speaker gender statistics. (\texttt{dev-real} demographics not included; see Section \ref{data-prep}.)}
  \label{tab:stats}
  \centering
  \begin{tabular}{p{20mm}p{10mm}p{10mm}p{10mm}p{15mm}}
    \toprule
    Split & Man & Woman & Non-Binary &
        (Prefer not to say)\\
    \midrule
    \texttt{train-real} & 54 &  17  & 0 & 3 \\
    \texttt{test-real}  & 5 & 4   &  1 & 0 \\
    \bottomrule
  \end{tabular}\label{stats-gender}
  
\end{table}

\begin{table}[ht]
  \caption{Speaker English proficiency statistics.}
  \centering
  \begin{tabular}{p{20mm}p{11mm}p{11mm}p{11mm}p{15mm}}
    \toprule
    Split & Native speaker & Fluent & Somewhat fluent &
        (Prefer not to say)\\
    \midrule
    \texttt{train-real} & 20 &  42  &  9 & 3 \\
    \texttt{test-real}  & 4  & 2  &  4 & 0 \\
    \bottomrule
  \end{tabular}\label{stats-fluency}
  
\end{table}

\begin{table}[ht]
  \caption{Speaker age ranges. (See \texttt{train-demographics.csv} and \texttt{test-demographics.csv} for more granularity.)}
  \centering
  \begin{tabular}{p{20mm}p{5mm}p{5mm}p{5mm}p{5mm}p{15mm}}
    \toprule
    Split & 18-25 & 26-35 & 36-45 & 46+ &
        (Prefer not to say)\\
    \midrule
    \texttt{train-real} & 11 & 41 & 6 &  1 & 15 \\
    \texttt{test-real}  & 3 & 5 & 2 & 0 & 0\\
    \bottomrule
  \end{tabular}\label{stats-age}
  
\end{table}

\section{Baseline models}

Here we describe extensive experiments with a set of baseline neural network models for Timers and Such. All experiments are conducted using the open source SpeechBrain \cite{SB2021} toolkit.

\subsection{ASR model and language models}\label{ASR}
The baseline models use an ASR model trained on the 960-hour LibriSpeech English ASR dataset \cite{librispeech}. The ASR model is an autoregressive attention-based sequence-to-sequence model \cite{bahdanau2014neural, chan2016listen} that achieves 3.08\% WER on the \texttt{test-clean} subset of LibriSpeech. The encoder of the ASR model extracts 40-dimensional FBANK features from the input signal and has two 2-D convolutional layers that downsample the input sequence by a factor of 4 in the time dimension, followed by four bidirectional LSTM layers and two fully-connected layers. The decoder is a GRU network that uses the location-aware attention mechanism of \cite{chorowski2015attention} to process the encoder outputs. The encoder outputs are additionally passed through a linear CTC \cite{Graves2006} head; during training, the output of the CTC head is used to compute an auxiliary CTC loss term \cite{kim2017joint}. Both the CTC head and the autoregressive decoder have 1000 outputs for a 1000-token SentencePiece \cite{kudo2018sentencepiece} BPE vocabulary.\footnote{More detailed hyperparameters for the ASR model can be found at \url{https://github.com/speechbrain/speechbrain/blob/develop/recipes/LibriSpeech/ASR/seq2seq/hparams/train_BPE_1000.yaml}.} (This ASR model was chosen because it was the best performing English ASR model in SpeechBrain at the time when these experiments were conducted.)

The ASR model transcribes the input signal $\mathbf{x}$ using a beam search for
\begin{align*}
    \argmax_{\mathbf{y}} & \phantom{+}\log p_{\text{ASR}}(\mathbf{y}|\mathbf{x})\\[-0.75em]
    &+ \alpha \log p_{\text{CTC}}(\mathbf{y}|\mathbf{x})\\
    &+ \beta \log p_{\text{LM}}(\mathbf{y})\\
    &+ \gamma c(\mathbf{x}, \mathbf{y}),
\end{align*}

where $p_{\text{CTC}}(\mathbf{y}|\mathbf{x})$ is the likelihood of transcript $\mathbf{y}$  according to the CTC head \cite{kim2017joint},
$p_{\text{LM}}(\mathbf{y})$ is the likelihood according to an external language model (LM), $c(\mathbf{x}, \mathbf{y})$ is a coverage penalty term \cite{kannan2018analysis}, and $\alpha, \beta, \gamma$ were set to minimize WER on the LibriSpeech dev sets.

The default LM is an LSTM trained on the LibriSpeech language modeling resources.\footnote{\url{https://www.openslr.org/11/}} In addition to the default LibriSpeech LM (LS LM), we also trained an LSTM LM on the Timers and Such training set transcripts (TAS LM). For ASR-based baseline models, we present results both using the LS LM and TAS LM.

\subsection{SLU models}\label{models}

We provide code, pre-trained models, and results for a traditional decoupled SLU model and (using the terminology suggested by Haghani et al. in \cite{Haghani2018}) two types of ``end-to-end'' models: a multistage model and a direct model.

The \textbf{decoupled} model uses a sequence-to-sequence model to map the transcript to the semantics. During training (and when decoding the validation set), the ground-truth transcripts are used as the input, and during testing, the transcripts produced by the LibriSpeech ASR model are used. For all models, the semantic dictionaries are treated as raw sequences of characters and split using a 51-token SentencePiece tokenizer.  

The \textbf{multistage} model likewise uses a sequence-to-sequence model to map the transcript to the semantics, but instead of training on the ground-truth transcripts, it is trained on the ASR transcripts. The transcripts are not precomputed: rather, each minibatch of audio signals is transcribed on the fly during training, which simplifies the implementation of our experiments.
In theory, transcribing training examples on the fly should also make the NLU model more robust, as it is exposed to more types of transcription errors resulting from different noise samples (e.g.~from dropout, batch normalization, data augmentation) across minibatches---though we have not compared the results with simply training on a single set of precomputed ASR transcripts, and leave this as an avenue for other researchers to explore. The downside of on-the-fly transcription is that the inherently sequential ASR beam search becomes a bottleneck on training step time. Using the default ASR beam width of 80, the time for one epoch on \texttt{train-synth} was about 12 hours (compared with about 0.5 hours for the decoupled model). Reducing the ASR beam width to 1 reduced the time for one epoch to about 2.5 hours. The results presented below use an ASR beam width of 1 for the multistage model.

The \textbf{direct} model uses a single sequence-to-sequence model to map audio directly to semantics, without an intermediate ASR search step. 
Compared to the multistage model, the direct model is significantly faster both in training and decoding, at about 1.5 hours per epoch with \texttt{train-synth} instead of 2.5 hours. 
Pre-training using related ASR or NLU tasks has consistently been found to improve the performance of direct models \cite{lugosch2019speech, wang2020large, huang2020leveraging, sharma2021leveraging, chung2021splat}, so we pre-train the encoder here as well. 
In our experiments described in previous papers, the encoder of the direct model was pre-trained using force-aligned phoneme and word labels \cite{lugosch2019speech, lugosch2020using}. The pre-training strategy used in this paper is somewhat simpler: we extract the encoder from the LibriSpeech ASR model and use it as a feature extractor in the direct SLU model. 
Another difference is that we do not backpropagate into the pre-trained encoder and leave its weights frozen, which greatly reduces training time and memory consumption. A more thorough ablation study and comparison of pre-training strategies would be worthwhile to conduct, but we leave that for the future, since the point here is just to establish some reasonable baseline models for this dataset.

While the SLU models do use a beam search to produce the output sequence, there are a number of differences between the SLU decoder and the ASR decoder. The SLU beam search does not use a coverage penalty (which was found to hurt performance both for Timers and Such and for the SLURP dataset) or an external ``language model'' over the space of output dictionaries. Instead of location-aware attention (which assumes a monotonic alignment between input and output sequences), the SLU decoder uses a simple one-headed key-value attention mechanism. The SLU models also do not use an auxiliary CTC head: whereas CTC's assumptions (monotonic alignments; output length $<$ input length) make sense for ASR, they generally do not hold for SLU, unless the dataset has word-aligned slot labels (Timers and Such does not). Other hyperparameters for these models were not optimized and chosen simply by copying the decoder hyperparameters from the LibriSpeech recipe, which were optimized for the validation set of that dataset.

\subsection{Experiments}\label{experiments}

For all baseline models, we provide results for three composite training sets: \texttt{train-real} only (trained for 50 epochs), \texttt{train-real} plus \texttt{train-synth} (trained for 2 epochs), and \texttt{train-synth} only (trained for 2 epochs). For all three training sets, we measure performance on \texttt{test-real} and \texttt{test-synth}. When training on \texttt{train-synth} only, we additionally report performance for \texttt{all-real}, a subset obtained by combining all the real data in \texttt{train-real}, \texttt{dev-real}, and \texttt{test-real}. (We do not test models trained on \texttt{train-real} on \texttt{all-real} because \texttt{all-real} contains \texttt{train-real}. For the same reason, we use \texttt{dev-synth}, not \texttt{dev-real}, to select the model checkpoint from the epoch with the best validation performance when testing on \texttt{all-real}.)

As in previous work, we report performance in terms of accuracy, where an output is deemed ``correct'' if all predicted slots and slot values are correct. Bastianelli et al. in \cite{slurp} have argued for the use of metrics more informative than simple accuracy when evaluating end-to-end SLU models. They propose SLU-F1, a metric based on word-level and character-level edit distance between the model's output and the true labels. The SLU-F1 metric sensibly penalizes errors like ``pizzas'' $\rightarrow$ ``pizza'' less than errors like ``pizzas'' $\rightarrow$ ``fries''. It is unclear, though, whether character-level edit distance is suitable for the numeric commands of Timers and Such: should ``11'' $\rightarrow$ ``111'' (character error rate of 50\%) be regarded as less of an error than ``11'' $\rightarrow$ ``22'' (character error rate of 100\%) when setting a cooking timer in minutes? For this reason, we do not recommend using character-level error to evaluate systems for this task. As a compromise, we also suggest reporting ``SLU WER'', an easy-to-compute metric that treats the space-delimited output of the SLU model and the true output dictionary as regular sequences of words and simply computes the usual WER metric. Note that no ``normalization'' of the outputs (e.g., "twelve and a half", "twelve point five" $\to$ ``12.5'') is necessary before evaluating, since the labels are always written in the correct numeric format.

\subsection{Results}\label{results-text}

\begin{table*}[]
\centering
\caption{Results (mean and stdev.~over 5 random seeds) for all baseline models. See Sec.~\ref{experiments} for the definition of ``SLU WER''.}
\label{comparison-all}
\begin{tabular}{cccccc}
\toprule
& & \multicolumn{2}{c}{\texttt{test-real}}  & \multicolumn{2}{c}{\texttt{test-synth}}              \\
    \cmidrule(r){3-4} \cmidrule(r){5-6}
\thead{Model} & \thead{Training set} & \thead{Accuracy} & \thead{SLU WER} & \thead{Accuracy} & \thead{SLU WER} \\

\midrule
\multirow{3}{*}{\parbox{1.5cm}{\centering Decoupled (LS LM)}} & \texttt{train-real}
 & $24.1\%_{\pm 1.1\%}$ & $34.4\%_{\pm 3.3\%}$ & $16.1\%_{\pm 1.4\%}$ & $33.2\%_{\pm 8.7\%}$ \\
 & (both)
 & $31.4\%_{\pm 4.3\%}$ & $26.5\%_{\pm 5.0\%}$ & $22.5\%_{\pm 2.1\%}$ & $25.2\%_{\pm 2.5\%}$  \\
 & \texttt{train-synth}
 & $32.3\%_{\pm 3.9\%}$ & $26.5\%_{\pm 2.5\%}$ & $23.7\%_{\pm 1.6\%}$ & $24.2\%_{\pm 0.7\%}$ \\
\midrule
\multirow{3}{*}{\parbox{1.5cm}{\centering Decoupled (TAS LM)}} & \texttt{train-real}
 & $43.5\%_{\pm 2.0\%}$ & $20.3\%_{\pm 3.5\%}$ & $34.6\%_{\pm 1.2\%}$ & $18.5\%_{\pm 3.8\%}$ \\
 & (both)
 & $46.8\%_{\pm 2.1\%}$ & $16.5\%_{\pm 2.2\%}$ & $38.4\%_{\pm 1.3\%}$ & $15.2\%_{\pm 0.9\%}$ \\
 & \texttt{train-synth}
 & $49.1\%_{\pm 2.3\%}$ & $16.3\%_{\pm 1.1\%}$ & $39.9\%_{\pm 0.7\%}$ & $13.9\%_{\pm 0.8\%}$ \\
\midrule
\multirow{3}{*}{\parbox{1.5cm}{\centering Multistage (LS LM)}} & \texttt{train-real}
 & $55.5\%_{\pm 3.4\%}$ & $10.1\%_{\pm 0.6\%}$ & $43.1\%_{\pm 2.9\%}$ & $10.8\%_{\pm 0.8\%}$ \\
 & (both)
 & $67.8\%_{\pm 1.4\%}$ & $7.4\%_{\pm 0.4\%}$ & $79.4\%_{\pm 0.4\%}$ & $3.2\%_{\pm 0.1\%}$ \\
 & \texttt{train-synth}
 & $66.6\%_{\pm 0.8\%}$ & $7.7\%_{\pm 0.8\%}$ & $79.1\%_{\pm 0.2\%}$ & $3.2\%_{\pm 0.0\%}$ \\
\midrule
\multirow{3}{*}{\parbox{1.5cm}{\centering Multistage (TAS LM)}} & \texttt{train-real}
 & $64.0\%_{\pm 3.3\%}$ & $7.4\%_{\pm 0.9\%}$ & $51.5\%_{\pm 2.9\%}$ & $8.7\%_{\pm 0.7\%}$ \\
 & (both)
 & $72.6\%_{\pm 1.6\%}$ & $5.9\%_{\pm 0.1\%}$ & $85.4\%_{\pm 0.2\%}$ & $2.4\%_{\pm 0.0\%}$ \\
 & \texttt{train-synth}
 & $72.2\%_{\pm 1.4\%}$ & $6.2\%_{\pm 0.4\%}$ & $85.4\%_{\pm 0.3\%}$ & $2.4\%_{\pm 0.1\%}$ \\
\midrule
\multirow{3}{*}{Direct} & \texttt{train-real}
 & $\mathbf{81.6}\%_{\pm 5.4\%}$ & $\mathbf{2.6}\%_{\pm 1.1\%}$ & $70.0\%_{\pm 5.7\%}$ & $15.2\%_{\pm 19.1\%}$ \\
 & (both)
 & $77.5\%_{\pm 1.6\%}$ & $3.3\%_{\pm 0.4\%}$ & $\mathbf{96.7}\%_{\pm 0.3\%}$ & $\mathbf{1.1}\%_{\pm 0.0\%}$ \\
 & \texttt{train-synth}
 & $68.0\%_{\pm 5.5\%}$ & $8.9\%_{\pm 3.4\%}$ & $96.4\%_{\pm 0.2\%}$ & $\mathbf{1.1}\%_{\pm 0.0\%}$  \\

\bottomrule
\end{tabular}
\label{results}
\end{table*}

\begin{table*}[]
\centering
\caption{Baseline results for the \texttt{all-real} set.}
\label{comparison-all-real}
\begin{tabular}{cccc}
\toprule
& & \multicolumn{2}{c}{\texttt{all-real}}                \\
    \cmidrule(r){3-4}
\thead{Model} & \thead{Training set}  & \thead{Accuracy} & \thead{SLU WER} \\

\midrule
Decoupled (LS LM) & \texttt{train-synth}
 & $26.8\%_{\pm 3.3\%}$ & $29.0\%_{\pm 2.2\%}$ \\
\midrule
Decoupled (TAS LM) & \texttt{train-synth}
 &  $44.6\%_{\pm 2.4\%}$ & $17.3\%_{\pm 1.1\%}$ \\
\midrule
Multistage (LS LM) & \texttt{train-synth}
 &  $64.6\%_{\pm 0.7\%}$ & $7.2\%_{\pm 0.2\%}$ \\
\midrule
Multistage (TAS LM) & \texttt{train-synth}
 &  $\mathbf{69.9}\%_{\pm 0.9\%}$ & $\mathbf{6.0}\%_{\pm 0.2\%}$ \\
\midrule
Direct & \texttt{train-synth}
 & $68.9\%_{\pm 5.4\%}$ & $8.2\%_{\pm 3.4\%}$ \\

\bottomrule
\end{tabular}
\label{results-all-real}
\end{table*}

A few trends in the results shown in Table \ref{results} are worth noting.

\begin{itemize}
\item \textbf{The direct model and multistage TAS LM work best.} This is perhaps unsurprising, since these two models effectively have the most opportunity to train on the downstream SLU task. 

% \item \textbf{The direct model can overfit to synthetic speech.} It seems that because the direct model has access to the raw speech features instead of a transcript, it can learn the idiosyncratic pronunciations of the speech synthesizer and achieve much better performance than the ASR-based models (96.7\% vs. 85.4\%).

\item \textbf{The direct model ``overfits'' to synthetic speech.} It seems that because the direct model has access to the raw speech features instead of a transcript, it can learn the idiosyncratic pronunciations of the speech synthesizer and achieve much better performance than the ASR-based models (96.7\% vs. 85.4\%). This model still performs well on the real test data---we mention this simply to explain why this model suddenly performs so much better for the synthetic test data.

\item \textbf{Test accuracies and SLU WERs\footnote{The 19.1\% stdev.~in SLU WER for the direct model on \texttt{test-synth} is due to a single outlier random seed for which the decoder produced many infinitely looping outputs (``\texttt{unit1 unit1 unit1 unit1}\dots'').} have high variability.} Some test accuracies have a standard deviation as high as 5.7\%. We observed this phenomenon with the preliminary version of Timers and Such and suspected that the variance was because of the smaller test set size (55 audios). However, this does not seem to be the explanation here, since \texttt{all-real} (Table \ref{results-all-real}) has 2,151 audios and still has highly variable test accuracy (stdev.~of 3.3\%, 2.4\%, 0.7\%, 0.9\%, 5.4\%). 
% Fig.~\ref{fig:valid-plot} shows validation set accuracy over time, which is correlated with the final test accuracy. For a given seed, there is not much variability over time at convergence, but the accuracy to which each seed converges varies greatly. 
We will not venture further here to diagnose this problem; instead, we leave it as a problem for future research on this dataset to solve.
\end{itemize}

\subsection{Computing resource usage}\label{compute-usage}
Training and testing all the SLU models across all random seeds, models, and training set compositions required about  233 GPU-hours on an Nvidia Quadro RTX 8000 GPU.
% (9 hours $\times$ 5 seeds for the direct model, 14.5 hours $\times$ 5 seeds for the multistage model for both the LS LM and TAS LM, and 2.5 hours $\times$ 5 seeds $+$ (3 hours $\times$ 5 seeds) $\times$ 2 for the decoupled model, which we trained once and subsequently decoded using both LMs). 
Additionally, the LibriSpeech ASR model was trained using one Nvidia Tesla V100 GPU for 194 hours, and the LibriSpeech LM was trained using 4 V100s for about 84 hours.

However, we hasten to note for those with limited computing resources interested in experimenting with Timers and Such that i) the pre-trained LibriSpeech models are available online and are downloaded automatically by the recipes, and ii) training a \textit{single} model on Timers and Such can be done relatively quickly, at around a minute per epoch for the direct recipe when training on \texttt{train-real}. The decoupled recipe can also be sped up significantly by using a larger batch size during training, since the input is text instead of speech and requires less memory. Note also that all the recipes have also been successfully tested on an older 12 GB Nvidia Tesla K80 GPU without any hyperparameter modifications.

\section{Potential social impact}\label{social-impact}
A risk of recording speech data is that a malicious actor could use the data to imitate the speaker and use the speaker's voice for purposes the speaker did not intend \cite{ovadya2019reducing}. Similar to Common Voice, it is unlikely that this could happen to the speakers of Timers and Such, since they did not provide any information that could uniquely identify them. 

On the whole, we think Timers and Such will be a great benefit to the research community and (indirectly) to users of voice interfaces. Speech datasets are often recorded by professional speakers in clean conditions unlike the conditions in which voice interfaces are typically used. This leads to brittle, overfitted models that break when applied to real-world speech \cite{likhomanenko2020rethinking}. Timers and Such will contribute to research and development of more robust models that can understand speech in a variety of accents and conditions.

\section{Conclusion}

Timers and Such is a new dataset of numeric commands that should be useful for SLU researchers, hackers aiming to train their own offline voice assistant, and researchers developing new representation learning methods for audio and language \cite{tamkin2020viewmaker, chung2020semi, pascual2019learning, baevski2020wav2vec} looking for another downstream task to test on. Some directions for the future of Timers and Such we hope to see worked on include: diagnosing and fixing the high variability of test performance; exploring the ASR model architecture (e.g., using a CTC model or transducer model \cite{graves2012sequence}); speeding up the multistage approach, e.g. by using transfer learning to initialize a multistage model using a decoupled model;  improving the performance of the direct model on \texttt{all-real}; using an ASR dataset with a more diverse set of accents and recording conditions, like Common Voice \cite{ardila2019common}; using different tokenizers or other hand-crafted output labels; improving the speech synthesis (using systems such as the RTVC multispeaker TTS  \cite{jia2018transfer, jemine2019master} to add even more synthetic speakers) and balance between real and synthetic training data; and enabling streaming inference \cite{mhiri2020low, potdar2021streaming}, which cannot be performed with the baseline models as-is, due to their global attention mechanism.

\section*{Acknowledgments}
Thanks to Olexa Bilaniuk for help with using the Mila cluster, Ju-Chieh Chou and Brian S.~Yeh for writing the LibriSpeech recipes and beam search utilities, and Aku Rouhe and Peter Plantinga for designing and implementing many nice features of SpeechBrain that made the experiments for this paper a lot easier to run. 

Timers and Such would not have been possible without the speakers who kindly took the time to donate their voices to the dataset and the friends who shared the project advertisement on social media.

\bibliographystyle{IEEEtran}

\bibliography{mybib}

\begin{thebibliography}{10}
\providecommand{\url}[1]{#1}
\csname url@samestyle\endcsname
\providecommand{\newblock}{\relax}
\providecommand{\bibinfo}[2]{#2}
\providecommand{\BIBentrySTDinterwordspacing}{\spaceskip=0pt\relax}
\providecommand{\BIBentryALTinterwordstretchfactor}{4}
\providecommand{\BIBentryALTinterwordspacing}{\spaceskip=\fontdimen2\font plus
\BIBentryALTinterwordstretchfactor\fontdimen3\font minus
  \fontdimen4\font\relax}
\providecommand{\BIBforeignlanguage}[2]{{%
\expandafter\ifx\csname l@#1\endcsname\relax
\typeout{** WARNING: IEEEtran.bst: No hyphenation pattern has been}%
\typeout{** loaded for the language `#1'. Using the pattern for}%
\typeout{** the default language instead.}%
\else
\language=\csname l@#1\endcsname
\fi
#2}}
\providecommand{\BIBdecl}{\relax}
\BIBdecl

\bibitem{hakkani2006beyond}
D.~Hakkani-T{\"u}r, F.~B{\'e}chet, G.~Riccardi, and G.~Tur, ``Beyond {ASR}
  1-best: Using word confusion networks in spoken language understanding,''
  \emph{Computer Speech \& Language}, vol.~20, no.~4, pp. 495--514, 2006.

\bibitem{saade2018spoken}
A.~Saade, A.~Coucke, A.~Caulier, J.~Dureau, A.~Ball, T.~Bluche, D.~Leroy,
  C.~Doumouro, T.~Gisselbrecht, F.~Caltagirone, T.~Lavril, and M.~Primet,
  ``Spoken language understanding on the edge,'' \emph{NeurIPS Workshop on
  Energy Efficient Machine Learning and Cognitive Computing}, 2019.

\bibitem{Serdyuk2018}
D.~Serdyuk, Y.~Wang, C.~Fuegen, A.~Kumar, B.~Liu, and Y.~Bengio, ``{Towards
  end-to-end spoken language understanding},'' \emph{ICASSP}, 2018.

\bibitem{lugosch2019speech}
L.~Lugosch, M.~Ravanelli, P.~Ignoto, V.~S. Tomar, and Y.~Bengio, ``Speech model
  pre-training for end-to-end spoken language understanding,''
  \emph{Interspeech}, 2019.

\bibitem{slurp}
E.~Bastianelli, A.~Vanzo, P.~Swietojanski, and V.~Rieser, ``{SLURP: A Spoken
  Language Understanding Resource Package},'' in \emph{{Proceedings of the 2020
  Conference on Empirical Methods in Natural Language Processing (EMNLP)}},
  2020.

\bibitem{leonard1993tidigits}
R.~G. Leonard and G.~Doddington, ``Tidigits ldc93s10,'' \emph{Web Download.
  Philadelphia: Linguistic Data Consortium}, 1993.

\bibitem{warden2018speech}
P.~Warden, ``Speech commands: A dataset for limited-vocabulary speech
  recognition,'' \emph{arXiv preprint arXiv:1804.03209}, 2018.

\bibitem{zohar_jackson_2018_1342401}
\BIBentryALTinterwordspacing
Z.~Jackson, C.~Souza, J.~Flaks, Y.~Pan, H.~Nicolas, and A.~Thite,
  ``Jakobovski/free-spoken-digit-dataset: v1.0.8,'' 2018. [Online]. Available:
  \url{https://doi.org/10.5281/zenodo.1342401}
\BIBentrySTDinterwordspacing

\bibitem{tamkin2020viewmaker}
A.~Tamkin, M.~Wu, and N.~Goodman, ``Viewmaker networks: Learning views for
  unsupervised representation learning,'' \emph{ICLR}, 2021.

\bibitem{chung2020semi}
Y.-A. Chung, C.~Zhu, and M.~Zeng, ``Semi-supervised speech-language joint
  pre-training for spoken language understanding,'' \emph{NAACL}, 2021.

\bibitem{pascual2019learning}
S.~Pascual, M.~Ravanelli, J.~Serra, A.~Bonafonte, and Y.~Bengio, ``Learning
  problem-agnostic speech representations from multiple self-supervised
  tasks,'' \emph{Interspeech}, 2019.

\bibitem{baevski2020wav2vec}
A.~Baevski, H.~Zhou, A.~Mohamed, and M.~Auli, ``wav2vec 2.0: A framework for
  self-supervised learning of speech representations,'' \emph{NeurIPS}, 2020.

\bibitem{ardila2019common}
R.~Ardila, M.~Branson, K.~Davis, M.~Henretty, M.~Kohler, J.~Meyer, R.~Morais,
  L.~Saunders, F.~M. Tyers, and G.~Weber, ``{Common Voice}: A
  massively-multilingual speech corpus,'' \emph{LREC}, 2020.

\bibitem{lugosch2020using}
L.~Lugosch, B.~H. Meyer, D.~Nowrouzezahrai, and M.~Ravanelli, ``Using speech
  synthesis to train end-to-end spoken language understanding models,'' in
  \emph{IEEE International Conference on Acoustics, Speech and Signal
  Processing (ICASSP)}.\hskip 1em plus 0.5em minus 0.4em\relax IEEE, 2020, pp.
  8499--8503.

\bibitem{taigman2017voice}
Y.~Taigman, L.~Wolf, A.~Polyak, and E.~Nachmani, ``{VoiceLoop}: Voice fitting
  and synthesis via a phonological loop,'' \emph{ICLR}, 2018.

\bibitem{veaux2017cstr}
C.~Veaux, J.~Yamagishi, K.~MacDonald \emph{et~al.}, ``{CSTR VCTK corpus:
  English multi-speaker corpus for CSTR voice cloning toolkit},''
  \emph{University of Edinburgh. The Centre for Speech Technology Research
  (CSTR)}, 2017.

\bibitem{timit}
J.~S. Garofolo, L.~F. Lamel, W.~M. Fisher, J.~G. Fiscus, D.~S. Pallett, and
  N.~L. Dahlgren, ``{DARPA TIMIT Acoustic Phonetic Continuous Speech Corpus
  CDROM},'' 1993.

\bibitem{seo2021integration}
S.~Seo, D.~Kwak, and B.~Lee, ``Integration of pre-trained networks with
  continuous token interface for end-to-end spoken language understanding,''
  \emph{arXiv:2104.07253}, 2021.

\bibitem{SB2021}
M.~Ravanelli, T.~Parcollet, P.~Plantinga, A.~Rouhe, S.~Cornell, L.~Lugosch,
  C.~Subakan, N.~Dawalatabad, A.~Heba, J.~Zhong, J.-C. Chou, S.-L. Yeh, S.-W.
  Fu, C.-F. Liao, E.~Rastorgueva, F.~Grondin, W.~Aris, H.~Na, Y.~Gao, R.~D.
  Mori, and Y.~Bengio, ``{SpeechBrain}: A general-purpose speech toolkit,''
  2021, arXiv:2106.04624.

\bibitem{librispeech}
V.~Panayotov, G.~Chen, D.~Povey, and S.~Khudanpur, ``{LibriSpeech: an ASR
  corpus based on public domain audio books},'' \emph{ICASSP}, 2015.

\bibitem{bahdanau2014neural}
D.~Bahdanau, K.~Cho, and Y.~Bengio, ``Neural machine translation by jointly
  learning to align and translate,'' \emph{ICLR}, 2015.

\bibitem{chan2016listen}
W.~Chan, N.~Jaitly, Q.~Le, and O.~Vinyals, ``Listen, attend and spell: A neural
  network for large vocabulary conversational speech recognition,'' in
  \emph{2016 IEEE International Conference on Acoustics, Speech and Signal
  Processing (ICASSP)}.\hskip 1em plus 0.5em minus 0.4em\relax IEEE, 2016, pp.
  4960--4964.

\bibitem{chorowski2015attention}
J.~Chorowski, D.~Bahdanau, D.~Serdyuk, K.~Cho, and Y.~Bengio, ``Attention-based
  models for speech recognition,'' \emph{NeurIPS}, 2015.

\bibitem{Graves2006}
A.~Graves, S.~Fern{\'a}ndez, F.~Gomez, and J.~Schmidhuber, ``{Connectionist
  Temporal Classification: Labelling Unsegmented Sequence Data with Recurrent
  Neural Networks},'' \emph{ICML}, 2006.

\bibitem{kim2017joint}
S.~Kim, T.~Hori, and S.~Watanabe, ``Joint {CTC}-attention based end-to-end
  speech recognition using multi-task learning,'' in \emph{2017 IEEE
  international conference on acoustics, speech and signal processing
  (ICASSP)}.\hskip 1em plus 0.5em minus 0.4em\relax IEEE, 2017, pp. 4835--4839.

\bibitem{kudo2018sentencepiece}
T.~Kudo and J.~Richardson, ``{SentencePiece}: A simple and language independent
  subword tokenizer and detokenizer for neural text processing,'' \emph{arXiv
  preprint arXiv:1808.06226}, 2018.

\bibitem{kannan2018analysis}
A.~Kannan, Y.~Wu, P.~Nguyen, T.~N. Sainath, Z.~Chen, and R.~Prabhavalkar, ``An
  analysis of incorporating an external language model into a
  sequence-to-sequence model,'' in \emph{2018 IEEE International Conference on
  Acoustics, Speech and Signal Processing (ICASSP)}.\hskip 1em plus 0.5em minus
  0.4em\relax IEEE, 2018, pp. 1--5828.

\bibitem{Haghani2018}
P.~Haghani, A.~Narayanan, M.~Bacchiani, G.~Chuang, N.~Gaur, P.~Moreno,
  R.~Prabhavalkar, Z.~Qu, and A.~Waters, ``{From Audio to Semantics: Approaches
  to end-to-end spoken language understanding},'' \emph{IEEE Spoken Language
  Technology Workshop (SLT)}, 2018.

\bibitem{wang2020large}
P.~Wang, L.~Wei, Y.~Cao, J.~Xie, and Z.~Nie, ``Large-scale unsupervised
  pre-training for end-to-end spoken language understanding,'' in \emph{ICASSP
  2020-2020 IEEE International Conference on Acoustics, Speech and Signal
  Processing (ICASSP)}.\hskip 1em plus 0.5em minus 0.4em\relax IEEE, 2020, pp.
  7999--8003.

\bibitem{huang2020leveraging}
Y.~Huang, H.-K. Kuo, S.~Thomas, Z.~Kons, K.~Audhkhasi, B.~Kingsbury, R.~Hoory,
  and M.~Picheny, ``Leveraging unpaired text data for training end-to-end
  speech-to-intent systems,'' in \emph{ICASSP 2020-2020 IEEE International
  Conference on Acoustics, Speech and Signal Processing (ICASSP)}.\hskip 1em
  plus 0.5em minus 0.4em\relax IEEE, 2020, pp. 7984--7988.

\bibitem{sharma2021leveraging}
B.~Sharma, M.~Madhavi, and H.~Li, ``Leveraging acoustic and linguistic
  embeddings from pretrained speech and language models for intent
  classification,'' in \emph{ICASSP 2021-2021 IEEE International Conference on
  Acoustics, Speech and Signal Processing (ICASSP)}.\hskip 1em plus 0.5em minus
  0.4em\relax IEEE, 2021, pp. 7498--7502.

\bibitem{chung2021splat}
Y.-A. Chung, C.~Zhu, and M.~Zeng, ``{SPLAT: Speech-Language Joint Pre-Training
  for Spoken Language Understanding},'' in \emph{Proceedings of the 2021
  Conference of the North American Chapter of the Association for Computational
  Linguistics: Human Language Technologies}, 2021, pp. 1897--1907.

\bibitem{ovadya2019reducing}
A.~Ovadya and J.~Whittlestone, ``Reducing malicious use of synthetic media
  research: Considerations and potential release practices for machine
  learning,'' \emph{arXiv preprint arXiv:1907.11274}, 2019.

\bibitem{likhomanenko2020rethinking}
T.~Likhomanenko, Q.~Xu, V.~Pratap, P.~Tomasello, J.~Kahn, G.~Avidov,
  R.~Collobert, and G.~Synnaeve, ``{Rethinking Evaluation in ASR: Are Our
  Models Robust Enough?}'' \emph{arXiv preprint arXiv:2010.11745}, 2020.

\bibitem{graves2012sequence}
A.~Graves, ``Sequence transduction with recurrent neural networks,'' \emph{ICML
  Workshop on Representation Learning}, 2012.

\bibitem{jia2018transfer}
Y.~Jia, Y.~Zhang, R.~J. Weiss, Q.~Wang, J.~Shen, F.~Ren, Z.~Chen, P.~Nguyen,
  R.~Pang, I.~L. Moreno \emph{et~al.}, ``Transfer learning from speaker
  verification to multispeaker text-to-speech synthesis,'' \emph{NeurIPS},
  2018.

\bibitem{jemine2019master}
C.~Jemine, ``Master thesis: Real-time voice cloning,'' 2019.

\bibitem{mhiri2020low}
M.~Mhiri, S.~Myer, and V.~S. Tomar, ``A low latency {ASR}-free end to end
  spoken language understanding system,'' \emph{Interspeech}, 2020.

\bibitem{potdar2021streaming}
N.~Potdar, A.~R. Avila, C.~Xing, D.~Wang, Y.~Cao, and X.~Chen, ``A streaming
  end-to-end framework for spoken language understanding,'' \emph{IJCAI}, 2021.

\end{thebibliography}

\appendix

\clearpage
\section{Appendix}

\subsection{Dataset documentation and intended uses}

The dataset is intended to be used for training, testing, and developing systems for speech recognition, spoken language understanding, and representation learning for audio and language. Other ``documentation'', such as demographics and steps taken during preparation, can be found in this paper in Section \ref{data-collection}.

\subsection{Consent form}\label{consent-form}
Participants were required to click a checkbox to the following consent form before recording (the page did not allow proceeding without clicking the checkbox):
    
\textit{Participating involves completing a voice sample recording process, which should take about 5 minutes. The recording process entails reading prompts that are displayed on the screen.}

                        \textit{Only your audio data, and whatever anonymous demographic information you choose to provide, will be stored on a web-based server and be made publicly available for download.}

                        \textit{\textbf{Risks.} A risk of making your voice data available is that someone could use your voice samples to imitate you. However, because your identity is not recorded or released, this risk is reduced.}

                        \textit{\textbf{Benefits.} Donating your voice samples does not guarantee you an immediate personal benefit, but indirectly you may benefit from the improved voice technology that will result from this project. The more diverse the set of voices we collect, the easier it will be for developers to make open source voice interfaces that work well for everyone.}

                        \textit{By continuing with this form, you acknowledge that you are at least 18 years old and agree to participate in the study.}
                        
                        \textit{I have read the description of the study and consent to participate in this study. <\textbf{checkbox}>}

\subsection{Particpant instructions}\label{participant-instructions}
Participants were given the following instructions before recording:

\textit{1. On the next page, you will be prompted for access to the microphone.}
    
\textit{2. You will see a prompt displayed on the screen. Click on the \textbf{red button} to record the prompt.}
    
\textit{3. Start speaking. When you have finished speaking, click on the \textbf{red button} again.}
    
\textit{4. You can listen to your recording by pressing the \textbf{Review} button.}
    
\textit{5. If you are not satisfied with your recording, click on the \textbf{Re-record} button.}
    
\textit{6. When you are satisfied with your recording, click on the \textbf{Next Prompt} button.}
    
\textit{7. Read the prompt casually, at a normal volume, as though you were talking to a human.}
    
\textit{8. \textbf{Do not refresh the page during your session!}}

\subsection{Data access}\label{data-access}

The dataset can be found at \url{https://zenodo.org/record/4623772}, where it can be easily downloaded either through the browser or using a command like \texttt{wget} \url{ https://zenodo.org/record/4623772/files/timers-and-such-v1.0.zip?download=1}. 

The DOI provided by Zenodo is 10.5281/zenodo.4623772.

Zenodo is a long-term research data storage platform; the dataset will continue to be hosted on Zenodo indefinitely. While we do hope to eventually expand Timers and Such with more speakers, intents, and languages other than English, the version described in this paper (Timers and Such v1.0) is static and is never intended to be changed. Thus, there is no planned maintenance or long-term preservation for the dataset, other than finding a new hosting solution in the event that Zenodo ceases to operate.

The code for the baseline model experiments can be found at \url{https://github.com/speechbrain/speechbrain/tree/develop/recipes/timers-and-such}. All experimental results can be obtained by running the command \texttt{run.sh} in the appropriate directory (``direct'', ``multistage'', ``decoupled'') on a machine with enough storage and a modern GPU.

A pre-trained model can be found at \url{https://huggingface.co/speechbrain/slu-timers-and-such-direct-librispeech-asr}. This model corresponds to the direct model with the best performing random seed for \texttt{test-real} among all the trials run.

The data is in the form of 16,000 Hz .wav files. Information and labels for each .wav file can be found in the \texttt{(train-real, dev-real, test-real, train-synth, dev-synth, test-synth).csv} files.

The easiest way to use the dataset is to run the SpeechBrain code provided above, which downloads the dataset and optionally prepares other useful information, like transcripts with normalized numeric values.

\subsection{Additional Information}

\begin{figure}[h]
    \centering
    \includegraphics[scale=0.8]{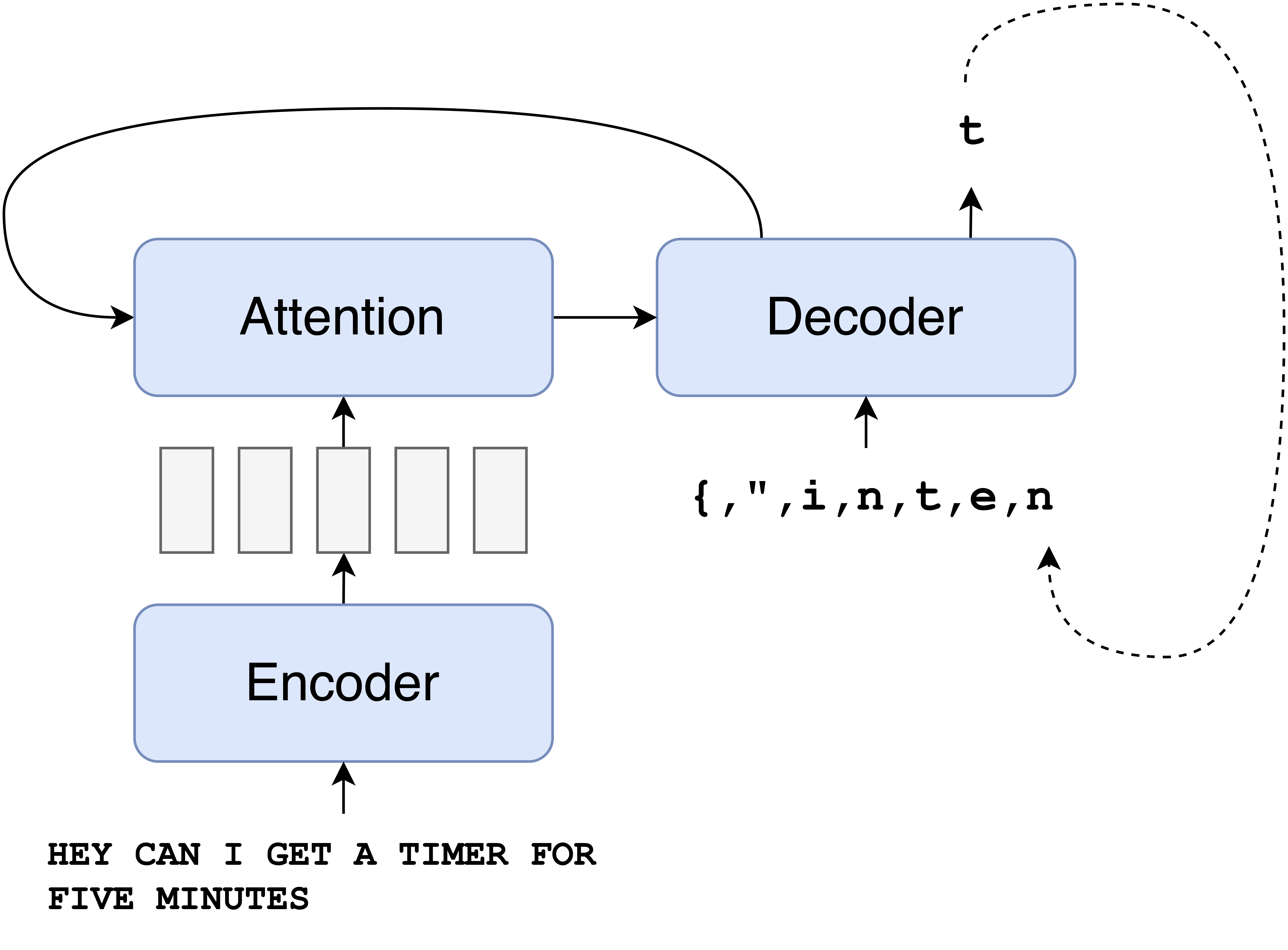}
    \caption{Diagram of the sequence-to-sequence model used in the transcript-based baselines.}
    \label{fig:model_diagram}
\end{figure}

Fig. \ref{fig:model_diagram} shows the architecture of the autoregressive sequence-to-sequence model used in our experiments. The input is the transcript (either the ground-truth transcript, for the decoupled model, or the ASR transcript, for the multistage model), and the output is the label dictionary, interpreted as a sequence of characters. The direct model is similar, except the input is speech instead of text.

\begin{table}[h]
  \caption{Size comparison of Timers and Such (TAS) with existing datasets.}
  \centering
  \begin{tabular}{p{20mm}p{10mm} p{10mm}p{10mm}p{10mm}p{10mm}p{10mm}} %{l r r r r r r}
    \toprule
     & FSC & Snips & SLURP (real) & SLURP (synth) & TAS (real) & TAS (synth) \\
    \midrule
    \# of speakers & 97 & 69  & 177 & 34 & 95 & 22 \\ 
    \# of utterances & 30,043 & 5,886 & 72,277 & 69,253 & 2,151 & 252,000 \\
    \# hours  & 19 & 5.5 & 58 & 43.5 & 2.5 & 171.5  \\
    \bottomrule
  \end{tabular}\label{stats-counts-compare}
  
\end{table}

Finally, Table \ref{stats-counts-compare} compares Timers and Such with existing open source SLU datasets. While the real subset of Timers and Such is relatively small, we emphasize that it is already large enough to train models that perform well on the real test set (see Table \ref{results}).

\end{document}